\documentclass[10pt,twocolumn,letterpaper]{article}
\usepackage{url}
\usepackage{stfloats}
\usepackage{cvpr}
\usepackage{times}
\usepackage{epsfig}
\usepackage{graphicx}
\usepackage{amsmath}
\usepackage{amssymb}
\usepackage{bm}
\usepackage[sort,numbers]{natbib}

\usepackage[utf8]{inputenc} 
\usepackage[T1]{fontenc}    
\usepackage{hyperref}       
\usepackage{url}            
\usepackage{booktabs}       
\usepackage{amsfonts}       
\usepackage{amsmath,amsthm,amssymb,amsfonts}
\usepackage{nicefrac}       
\usepackage{microtype}      
\usepackage{algorithm}  
\usepackage{algorithmic}
\usepackage{graphicx}
\usepackage{natbib}

\title{TGAN: Deep Tensor Generative Adversarial Nets for Large Image Generation}

\author{Zihan Ding\\
Imperial College London\\
{\tt\small zd2418@ic.ac.uk}
\and
Xiao-Yang Liu\\
Columbia University\\
{\tt\small xl2427@columbia.edu}
\and
Miao Yin\\
University of Electronic Science\\
and Technology of China\\
{\tt\small yinmiaothink@gmail.com}\\
\and 
Linghe Kong\\
Shanghai Jiao Tong University\\
{\tt\small linghe.kong@sjtu.edu.cn}\\
}

\begin{document}

\maketitle

\begin{abstract}

  Deep generative models have been successfully applied to many applications. However, existing works experience limitations when generating large images (the literature usually generates small images, e.g. $32 \times 32$ or $128 \times 128$). In this paper, we propose a novel scheme, called deep tensor adversarial generative nets (TGAN), that generates large high-quality images by exploring tensor structures. Essentially, the adversarial process of TGAN takes place in a tensor space. First, we impose tensor structures for concise image representation, which is superior in capturing the pixel proximity information and the spatial patterns of elementary objects in images, over the vectorization preprocess in existing works. Secondly, we propose TGAN that integrates deep convolutional generative adversarial networks and tensor super-resolution in a cascading manner, to generate high-quality images from random distributions.  More specifically, we design a tensor super-resolution process that consists of tensor dictionary learning and tensor coefficients learning. Finally, on three datasets, the proposed TGAN generates images with more realistic textures, compared with state-of-the-art adversarial autoencoders. The size of the generated images is increased by over $8.5$ times, namely $374\times 374$ in PASCAL2. 

\end{abstract}

\section{Introduction}

With the great success in deep learning, the deep generative model have been investigated widely. The generative adversarial nets (GAN) \cite{goodfellow2014generative} based methods are applied in many interesting applications including image super-resolution \cite{ledig2016photo}, image-to-image translation \cite{zhu2017unpaired}\cite{isola2017image}, text-to-image translation \cite{xu2017attngan}, dialogues generation \cite{serban2017hierarchical}, etc. With the development of graphical technologies, the demand of higher resolution images has increased significantly. Moreover, generation of large high-resolution images remains a challenge.

However, existing GAN models experience limitations when generating large images. With the growing scale of images, vanilla GAN is hard to produce high-quality natural images because it is difficult for the generator and the discriminator to achieve optimality simultaneously. When processing high-dimensional images, the computation complexity and the training time increases significantly. The challenge is that the image has too many pixels and it is hard for a single generator $G$ to learn the empirical distribution. Therefore, the traditional GAN \cite{goodfellow2014generative} does not scale well for the generation of large images. The variations of GAN such as deep convolutional GAN (DCGAN) \cite{radford2015unsupervised}, super-resolution GAN (SRGAN) \cite{ledig2017photo}, Laplacian Pyramid GAN (LAPGAN) \cite{denton2015deep} and StackGAN \cite{zhang2017stackgan} are promising candidates for generative models in unsupervised learning. It is desirable to construct a generative model that efficiently processes data with large size and high dimensions. 

Traditional GAN-based methods operates in pixel space to generate images while tensor-based methods work in tensor space. Tensor representation \cite{kolda2009tensor} and its derivative methods such as tensor sparse coding~\cite{jiangfei2018AAAI} and tensor super-resolution have a more concise and efficient representation of images, especially for large images. They provide an alternative method for representing large images in the tensor space, instead of the traditional pixel space or frequency domain, which could benefit challenges of generating large-sized high-resolution images.

Large-sized or high-dimensional images can be realized in several possible ways. Super-resolution \cite{yang2010image} is one of the classic methods used to construct high-resolution images from low-resolution images for better human interpretation. The key idea to achieve super-resolution is to use the nonredundant information contained in multiple low-resolution images induced by the subpixel shifts between them. One recent popular scheme for image super-resolution is SRGAN \cite{ledig2017photo}, which combines GAN with deep transposed convolutional neural networks (CNNs) for generating high-resolution images from low-resolution ones. The generator in SRGAN is used for upsampling the low-resolution images to super-resolution images, which are distinguished from the original high-resolution images by the discriminator.  

Dictionary learning \cite{mairal2009online}\cite{mairal2009supervised} is another method to efficiently to process large-sized or high-dimensional data. Using dictionary learning, we try to find sparse representation of input image data, which corresponds to the sparse coding technology of images. Traditional sparse coding method encodes images in matrices, while tensor-based sparse coding \cite{jiangfei2018AAAI} is more flexible with larger representation space. Multi-dimensional tensor sparse coding uses t-linear combination to obtain a more concise and small dictionary for representing the images, and the corresponding coefficients have richer physical explanations than the traditional methods.
We apply the basic principles of super-resolution and tensor-based dictionary learning in our generative model.

For large-sized and high-dimensional images, the tensor representation is able to preserve the local proximity and capture the spatial patterns of elementary objects. Existing conventional sparse coding only captures linear correlations, which harms the spatial patterns of images. However, tensor sparse coding model can capture nonlinear correlations (linear upon sine/cosine basis), which is consistent with the existing neural networks using nonlinear activation functions. Tensor sparse coding replaces conventional vectorizing process with tensorizing process \cite{qi2016tensr}\cite{jiangfei2018AAAI}\cite{cohen2016expressive}\cite{sharir2016tractable}. For complex and high-dimensional images, the conventional sparse coding process uses vector representation, and the vectorizing process ignores the spatial structure of the data. As a result, it generates a large-sized dictionary and causes high-computational complexity, which makes it infeasible for high-dimensional data applications. 

Tensor-based dictionary learning adopts a series of dictionaries to approximate the structures of the input data in each scale, which significantly reduces the size of the dictionaries. Besides, the circular matrix defined at Section 3.1 maintains the original image invariant after shifting; this helps to preserve the spatial structure of the images. Benefitting from tensor representation, tensor-based dictionary learning has advantages in dictionary size, shifting invariance, and rich physical explanations of the tensor coefficients~\cite{jiangfei2018AAAI}. In general, tensor-based methods have a more efficient representation capability for large-sized or high-dimensional data, and could therefore benefit the generative models. We believe that incorporating the tensor-based methods includig tensor representation, tensor sparse coding, and tensor super-resolution in the generative models will improve large-sized high-resolution images generation. 

In this paper, we present a novel generative model called deep tensor generative adversarial nets (TGAN), cascading a DCGAN and tensor-based super-resolution to generate large-sized high-quality images (e.g. $374\times 374$). The contribution of the proposed TGAN is threefold: (i) We apply tensor representation and tensor sparse coding for images representation in generative models. This is testified to have advantages of more concise and efficient representation of images with less loss on spatial patterns. (ii) We incorporate the tensor representation into the super-resolution process, which is called tensor super-resolution. The tensor super-resolution is cascaded after a DCGAN with transposed convolutional layers, which generates low-resolution images directly from random distributions. (iii) The DCGAN and tensor dictionaries in tensor super-resolution are both pretrained with a large number of high-resolution and low-resolution images. The size of dictionaries is smaller with tensor representation than traditional, which accelerates the dictionary learning process in tensor super-resolution.  
More details are shown in Fig. 1 for an illustration of the TGAN. The generation performance of TGAN surpasses traditional generative models including adversarial autoencoders \cite{makhzani2015adversarial} in inception score \cite{salimans2016improved} on test datasets, especially for large images. Our code is available at https://github.com/hust512/Tensor-GAN. 



\section{Related Work}
\label{sect: Related Work}
Recently, various approaches have been developed to study the deep generative model. There are two main types of the generative models that includes the adversarial model GAN \cite{goodfellow2014generative} and its modifications, and the probability model such as variational autoencodes (VAE) \cite{kingma2013auto} and adversarial autoencoders (AAE) \cite{makhzani2015adversarial}.  

GAN is a two-player game that consists of a generator $G$ and a discriminator $D$. The generator $G$ can generate realistic samples based on the input random noise, while the discriminator $D$ is aimed to identify whether the samples come from the real sample set or the generated data set. Finally, $G$ and $D$ reach a Nash equilibrium and $G$ is able to generate stable images. However, large images make this equilibrium hard to reach for $G$ and $D$ at the same time.

In order to generate high-resolution images from low-resolution images, the model SRGAN \cite{ledig2017photo} is proposed to realize super-resolution of images. It uses CNN for extracting features from low-resolution images.
The model of SRGAN testifies the strong capability of generative models in applications of images super-resolution. Another popular and successful modification of the GAN is DCGAN \cite{radford2015unsupervised} comprising transposed CNNs, especially for images-related applications for unsupervised learning in computer vision. Convolutional strides and transposed convolution are applied for the downsampling and upsampling. 
However, even with DCGAN, the bottleneck of GAN could be achieved easily for large images, which is that increasing the complexity of the generator does not necessarily improve the image quality. Moreover, StackGAN \cite{zhang2017stackgan} uses a two-stage GAN to generate images of size $256\times 256$, which are relatively large images for state-of-art generative models.

AAE \cite{makhzani2015adversarial} is a combination of GANs and VAE. AAE utilities only half of the autoencoder to map the original data distribution $x$ into the latent variable distribution $z$; then, it uses an adversarial approach to optimize $z$. The data sample generation is different between AAE and GAN. The GAN compares the generated data distribution with real data distribution in the discriminator and adopts a stochastic gradient descent process to optimize the entire model. On the other hand, AAE uses the discriminator to distinguish the latent variable distribution $z$. The discrete data that cannot be processed by the GAN is mapped to the continuous data in $z$, which extends the range of the acceptable data.


However, image representation in pixel space may not be an efficient way as in the traditional GANs. Tensor representation based methods have been adopted recently. Recent papers \cite{tan2015tensor}\cite{jiangfei2018AAAI} apply tensor representation for dictionary learning with smaller dictionary size and better results than the traditional methods. Some theoretical analysis for tensor decomposition and its application are provided in \cite{kolda2009tensor} with details. Tensor decomposition lies in the core status of tensor-based methods, which provide an alternative representation mean for data such as large images. 




\section{Notations and Preliminaries}
\label{sect: Deep Tensor GAN}
\begin{figure*}
  \centering
 \includegraphics[height=11cm]{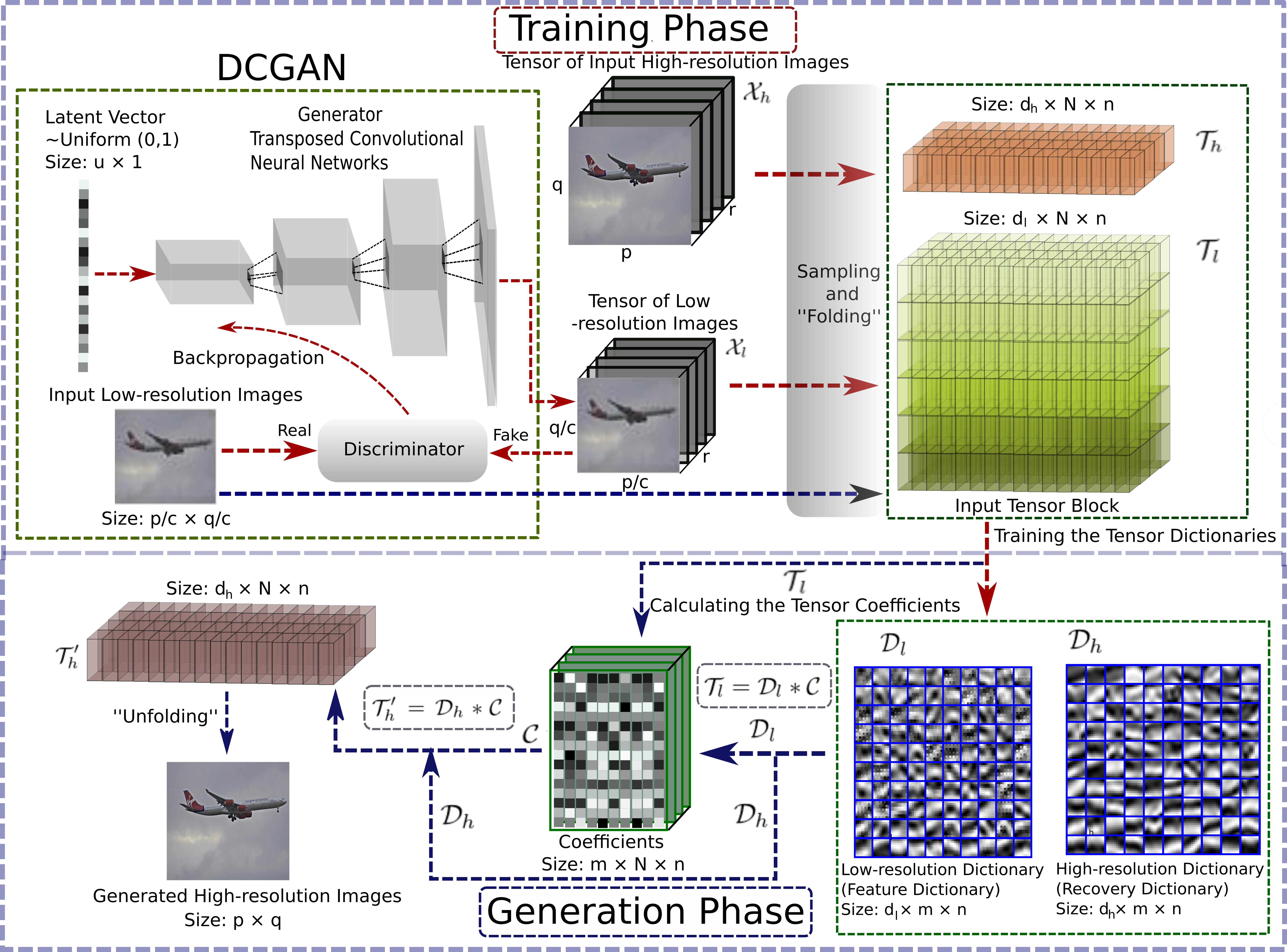}

  \caption{The architecture of TGAN. The latent vectors are sampled from random distributions. During the training phase, the DCGAN are trained with the input low-resolution images, to generate low-resolution image tensors from latent vectors. Through a sampling and ``folding'' process, the high-resolution and low-resolution image tensors are transformed into tensor blocks, $\mathcal{T}_h$ and $\mathcal{T}_l$, respectively. The feature dictionary (low-resolution) $\mathcal{D}_l$ and recovery dictionary (high-resolution) $\mathcal{D}_h$ are trained with these input tensor blocks. In the generation phase, low-resolution tensor images are generated with DCGAN from the latent vectors. The tensor coefficients $\mathcal{C}$ are obtained using $\mathcal{T}_l=\mathcal{D}_l*\mathcal{C}$, where $\mathcal{D}_l$ is the low-resolution tensor feature dictionary derived from the training phase. High-resolution tensor images can be obtained via $\mathcal{T}^\prime_h=\mathcal{D}_h*\mathcal{C}$, where $\mathcal{D}_h$ is the trained high-resolution tensor recovery dictionary. The final 2D images $X^\prime$ are transformed from the high-resolution tensor images $\mathcal{T}^\prime_h$. (Note that during the training phase $\mathcal{T}_l$ is derived from input low-resolution images while for the generation phase it is from images generated with DCGAN.)}
  \label{fig:TGAN}
\end{figure*}
\subsection{Tensor Product}
\label{sect: Notations}
We use boldface capital letters to denote matrices, e.g. $\bf{A}$, and calligraphic letters to denote tensors, e.g. $\mathcal{T}$. An order-3 tensor is denoted as $\mathcal{T} \in \mathbb{R}^{n_1 \times n_2 \times n_3}$. The expansion of $\mathcal{T}$ along the third dimension is represented as $\underline{\mathcal{T}} = [\mathcal{T}^{(1)}; \mathcal{T}^{(2)}; \cdot \cdot \cdot\mathcal{T}^{(k)};\cdot \cdot\cdot \mathcal{T}^{(n_3)}] \in \mathbb{R}^{n_1 n_2 \times n_3} $, where $\mathcal{T}^{(k)}$ denotes the $k$-th frontal slice, for $k=1,2,...,n_3$. The circular matrix representation of tensor $\mathcal{T}$ is defined as 
\begin{equation}
    \underline{\mathcal{T}}^{c} = \left[
 \begin{matrix}
   \mathcal{T}^{(1)} & \mathcal{T}^{(n_3)} & \cdot\cdot\cdot & \mathcal{T}^{(2)} \\
   \mathcal{T}^{(2)} & \mathcal{T}^{(1)} & \cdot\cdot\cdot & \cdot\cdot\cdot\\
   \cdot\cdot\cdot & \cdot\cdot\cdot & \cdot\cdot\cdot & \mathcal{T}^{(n_3)}  \\
   \mathcal{T}^{(n_3)} & \mathcal{T}^{(n_3 - 1)} & \cdot\cdot\cdot & \mathcal{T}^{(1)}
  \end{matrix}
  \right].
\end{equation}


The tensor product~\cite{hao2013facial} of two tensors $\mathcal{A} \in \mathbb{R}^{n_1 \times n_2 \times n_3}$ and $\mathcal{B} \in \mathbb{R}^{n_2 \times n_4 \times n_3}$ is defined as 
\begin{equation}
    \mathcal{T} = \mathcal{A} \ast \mathcal{B} \in \mathbb{R}^{n_1 \times n_4 \times n_3},
\end{equation} 
where $\mathcal{T}(i , j , :) = \sum^{n_2}_{s=1} \mathcal{A}(i , s, :) \ast \mathcal{B}(s , j , :)$ for $i = {1, 2,  ... , n_1}$ and  $j = {1, 2,  ... , n_4}$, and $\ast$ denotes the circular convolution operation. In addition, the tensor product has an equivalent matrix-product form:
\begin{equation}
    \underline{\mathcal{T}} = \underline{\mathcal{A}}^{c}\underline{\mathcal{B}}.
\end{equation}

\subsection{Tensor Sparse Coding for images}
\label{sect: Notations}
Considering $r$ input images $X$ of size $p\times q$, we first sample the image tensor $\mathcal{X}\in \mathbb{R}^{p\times q\times r}$ using tensor cubes and reshape it to be the input tensor block ${\mathcal{T}}\in \mathbb{R}^{d\times N \times n }$ (detailed relationships of $d,N,n$ with $p,q,r$ and the tensor cubes are shown in Section 4). ${\mathcal{T}}$ can be approximated with an overcomplete tensor dictionary $\mathcal{D}\in \mathbb{R}^{d \times m \times n}$, $m>d$ as follows~\cite{jiangfei2018AAAI}:
\begin{equation}
    {\mathcal{T}}=\mathcal{D}\ast\mathcal{C}={\mathcal{D}_1}\ast \mathcal{C}_1+...+{\mathcal{D}_m}\ast \mathcal{C}_m,
\end{equation}
where $\mathcal{C}\in \mathbb{R}^{m \times N \times n}$ is the tensor coefficient with slice $\mathcal{C}_j=\mathcal{C}(j,:,:)$.

One of the proposed schemes for tensor sparse coding is based on the $\ell_1$-norm of the coefficient. The sparse coding problem in tensor representation is as follows:
\begin{equation}
    \min_{\mathcal{D,C}}\quad \frac{1}{2}||\mathcal{X}-\mathcal{D}\ast\mathcal{C}||^2_F+\lambda||\mathcal{C}||_1
\end{equation}
\begin{equation}
    \textrm{s.t.} \quad ||\mathcal{D}(:,j,:)||^2_F\leqslant1, j=1,2...m,
\end{equation}
where the size of the dictionary $\mathcal{D}$ is $d \times m \times n $, $m>d$. However, traditional sparse coding requires the size of the dictionary to be $(d\times n)\times m, m>d\times n$, which significantly increases with the increase in dimensionality, as shown in~\cite{jiangfei2018AAAI}. A smaller dictionary is easier to learn in tensor sparse coding, which is a more efficient way to encode images compared with traditional sparse coding methods.

\section{Deep Tensor Generative Adversarial Nets Scheme}
We incorporate tensor-based methods including tensor representation, tensor sparse coding, tensor dictionary learning, and tensor super-resolution into traditional generative models such as DCGAN. 
The proposed novel scheme is called TGAN. 

The TGAN scheme could be divided into two phases: the training phase and the generation phase, as shown in Fig. 1. First of all, two-dimensional (2D) images are transformed into the tensor space as a preprocess. In the generation phase: using pretrained DCGAN to generate low-resolution image tensors from random distributions, we apply tensor super-resolution for transforming low-resolution image tensors to high-resolution image tensors. High-resolution 2D images can be derived from the obtained high-resolution image tensors. The tensor dictionaries we used in the tensor super-resolution process and the DCGAN are both pretrained with large numbers of high-resolution and low-resolution image tensors in the training phase. It is clear that the training phase is ahead of the generation phase in implementations. 

We sequentially introduce details of the TGAN scheme in the following subsections. Subsection 4.1 provides a basic introduction to tensor representation applied in our TGAN scheme. In Subsection 4.2, we propose the ``folding'' and ``unfolding'' process of data preparation for tensor dictionary learning. In Subsection 4.3, we present the training phase of TGAN scheme, including the DCGAN training and tensor dictionaries learning. Subsection 4.4 provides details about tensor super-resolution process, including theories and implementations. In Subsection 4.5, we present the generation phase of the TGAN scheme, which generates the super-resolution images with the trained DCGAN and tensor dictionaries.

\subsection{Tensor Representation in TGAN}
\label{sect: The Architecture of Deep Tensor GAN}

Our proposed approach combines DCGAN with tensor-based super-resolution, to directly generate high-resolution images. Considering the advantages of small dictionary size and invariance of shifting~\cite{jiangfei2018AAAI}, tensor sparse coding is the key point we want to apply in our model. We make the assumption~\cite{she2018data} that the inner patterns of images can be at least approximately sparsely represented with a learned dictionary. For tensor dictionary representation, $\mathcal{T}=\mathcal{D}\ast\mathcal{C}$, where $\mathcal{T}\in \mathbb{R}^{d\times N \times n}, \mathcal{D}\in \mathbb{R}^{d\times m\times n}, \mathcal{C}\in \mathbb{R}^{m\times N \times n} $.
Therefore, tensor representation of images is necessary, which acts as the main representation of images in our workflows. 

\subsection{Data Preprocess: ``Folding'' and ``Unfolding''}
We obtain the tensor input block $\mathcal{T}$ with original images $X\in \mathbb{R}^{p\times q}$ in the following manner, which we called the ``folding'' process.
We first concatenate $r$ images shifted from the same original image $X\in\mathbb{R}^{p\times q}$ for high-resolution or $X\in\mathbb{R}^{\frac{p}{c}\times \frac{q}{c}}$ for low-resolution (first upsampling it to be $X\in\mathbb{R}^{p\times q}$ in the generation phase) with different pixels to obtain the image representation tensor $\mathcal{X}\in \mathbb{R}^{p\times q\times r}$, as shown in Fig. 2. Then we sample $N_0$ image tensors $\mathcal{T}$ in all dimensions with the tensor block of size $a\times a \times a$ to obtain $N$ sample blocks, where $N=N_0\times (p-a+1)\times (q-a+1)\times(r-a+1)$. Therefore, the size of image representation tensor is $(a\times a \times a)\times (p-a+1)\times (q-a+1)\times(r-a+1) $. The tensor is reshaped to be input tensor blocks $\mathcal{T}\in \mathbb{R}^{d\times N \times n}$, where $d=a\times a,n= a $. For tensor dictionary learning process, the original images $X$ are 2D images from the training set; for the image generation process with trained dictionaries, the original image $X$ is generated with DCGAN from random distributions, and with $N_0=1$ in order to generate a single high-resolution image from scratch. As the tensor dictionary $\mathcal{D}\in \mathbb{R}^{d\times m\times n}$ is independent of the number of samples $N$, the dictionary iteratively trained with a large number $N$ of samples could naturally be used for generating a single high-resolution image.
\begin{figure}[t]
  \centering
  \includegraphics[height=5cm]{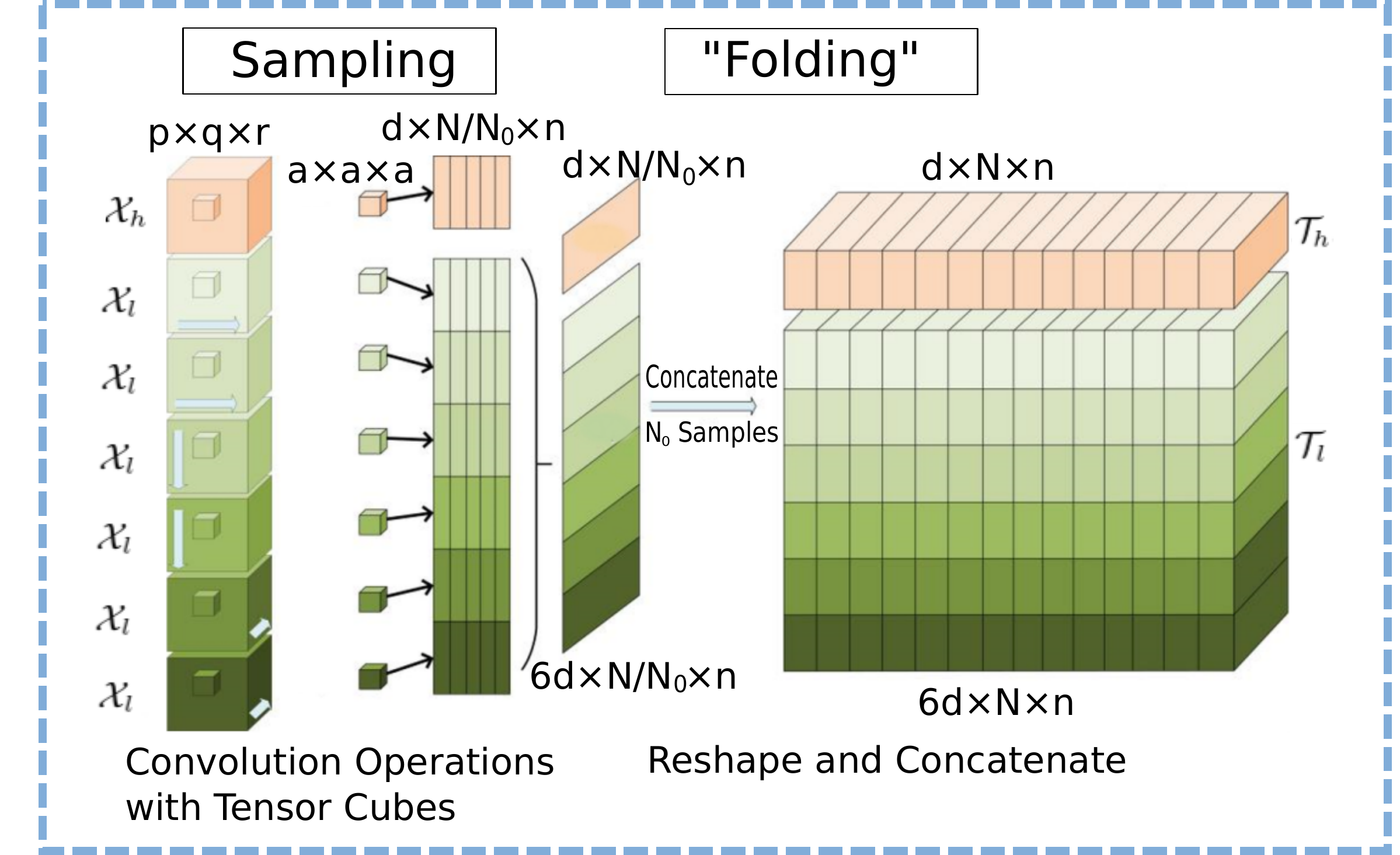}
  \caption{Preparation of the tensor blocks for tensor dictionary learning, including sampling and ``folding''. With the concatenated high-resolution image tensors $\mathcal{X}_h$ and low-resolution image tensors $\mathcal{X}_l$ (upsampled to have same size with $\mathcal{X}_h$) from the same original image, we sample (through a convolution operation) $\mathcal{X}_l$ in all dimensions and  $\mathcal{X}_h$ in one dimension with the tensor cubes of size $a\times a \times a$, to obtain $N$ sample blocks and reshape them. With a batch of original images, we could obtain the tensor blocks $\mathcal{T}_l\in\mathbb{R}^{d_l\times N\times n}$ and $\mathcal{T}_h\in\mathbb{R}^{d_h\times N\times n}$, where $d_l=6\times d, d_h=d$.}
  \label{fig:TGAN}
\end{figure}

The inverse process of the above ``folding'' process is called the ``unfolding'' process, which is used for recovering the high-resolution 2D images from the obtained high-resolution tensor output blocks. The ``unfolding'' is just a trivial combination of inversing each step in ``folding''.

\subsection{The Training Phase: DCGAN Training and Tensor Dictionary Learning}

In our model, we first downsample the original images $X\in \mathbb{R}^{p\times q}$ in the training set to high-resolution images $\bm{X}_h\in \mathbb{R}^{p\times q}$ and low-resolution images $\bm{X}_l\in \mathbb{R}^{\frac{p}{c}\times \frac{q}{c}}$ at the downsampling rate $c$, and we further transform them into tensor representation $\mathcal{X}_l, \mathcal{X}_h$. Then we train DCGAN with ${X}_l$ to generate low-resolution tensor images $\mathcal{T}_G \in \mathbb{R}^{\frac{p}{c} \times \frac{q}{c} \times r}$ from random distributions $\textbf{r} \sim $Uniform$(0,1)$. We refer to the adversarial loss as utilities. The reconstruction loss and adversarial loss is formulated as a minimax game:
\begin{align}
    \nonumber\min_{G}\max_{D} L(G,D) =& \mathbb{E}_{\textbf{r}\sim U} \left[\log(1 - D(G(\textbf{r})))\right]\\
    + &\mathbb{E} \left[\log{D({X}_l)}\right],
\end{align}
where $G,D$ denote generator and discriminator of DCGAN, and $\textbf{r},U$ denote the latent vector and uniform distributions. The images in tensor representation $\mathcal{X}_l$ and $\mathcal{X}_h$ are further transformed to be input tensor blocks $\mathcal{T}_l$ and $\mathcal{T}_h$ (as shown in the data preprocess of Section 4.2) for training the dictionaries $\mathcal{D}_l$ and $\mathcal{D}_h$ in tensor super-resolution. We have tensor product relationships in tensor sparse coding: $\mathcal{T}_h = \mathcal{D}_h * \mathcal{C}_h$ and $\mathcal{T}_l = \mathcal{D}_l * \mathcal{C}_l$, where $\mathcal{C}_h, \mathcal{C}_l$ denotes tensor sparse coefficients for high-resolution images and low-resolution images respectively. Note that, in tensor super-resolution, it is reasonable (reasons in Section 4.4) to set $\mathcal{C}_h=\mathcal{C}_l$ and denote it with $\mathcal{C}$.

\subsection{Details about Tensor Super Resolution}
The goal for tensor super-resolution is to transform low-resolution images ${X}_l$ into high-resolution images ${X}_h$ through the tensor spares coding approach. For an input tensor $\mathcal{T} \in \mathbb{R}^{d \times N \times n}$, tensor dictionary learning is similar to (the only difference is the dimensions) the tensor sparse coding in Section 3.2, 
where $\mathcal{D}\in\mathbb{R}^{d\times m \times n}$ is the tensor dictionary, and its slice $\mathcal{D}(:,j,:)$ is a basis, $\mathcal{C}\in \mathbb{R}^{m\times N\times n}$ is the tensor sparse coefficient. The first and second term uses the Frobenius norm and $\ell_1$-norm in Equ. (5), respectively.

If taking the sparse coding process of different resolution images as similar patterns with respect to different bases, we could consider that high-resolution and low-resolution tensor images from the same origins have sparse and approximate tensor coefficients $\mathcal{C}$. Therefore the constraints of two dictionaries could be combined as follows:
\begin{equation}
    \mathcal{D}=\arg\min_{\mathcal{D},\mathcal{C}}||\mathcal{X}-\mathcal{D}\ast\mathcal{C}||^2_F+\lambda||\mathcal{C}||_1,
\end{equation}
where
\begin{equation}
    \mathcal{X}=\begin{bmatrix}
    \frac{1}{\sqrt{N}}\mathcal{T}_h\\
    \frac{1}{\sqrt{M}}\mathcal{T}_l
    \end{bmatrix},
    \mathcal{D}=\begin{bmatrix}
    \frac{1}{\sqrt{N}}\mathcal{D}_h\\
    \frac{1}{\sqrt{M}}\mathcal{D}_l
    \end{bmatrix},
    \lambda=\frac{\lambda_h}{N}+\frac{\lambda_l}{M},
\end{equation}
where $\mathcal{T}_h,\mathcal{T}_l$ represent input tensor blocks of  high-resolution and low-resolution images and $N, M$ denote the number of samples in two kinds of resolutions. We then apply the Lagrange dual method and iterative shrinkage threshold algorithm based on tensor-product to solve the tensor dictionaries and tensor sparse coefficients.
The minimization problem can be rewritten as:
\begin{equation}
    \min_{\mathcal{C}}f(\mathcal{C})+\lambda g(\mathcal{C})
\end{equation}
where $f(\mathcal{C})$ stands for $\frac{1}{2}||\mathcal{X}-\mathcal{D}\ast\mathcal{C}||^2_F$ and $g(\mathcal{C})$ stands for $||\mathcal{Z}||_1$ (coefficient $\frac{1}{2}$ can be absorbed in $\lambda$). At the $(s+1)$-th iteration, 
    \begin{align}
    \nonumber\mathcal{C}_{s+1}& =\arg \min_{\mathcal{C}}f(\mathcal{C}_s)+\langle\nabla f(\mathcal{C}_s),\mathcal{C}-\mathcal{C}_s\rangle \\
    & +\frac{L_{s+1}}{2}||\mathcal{C}-\mathcal{C}_s||^2_F+\lambda g(\mathcal{C}),
    \end{align}
where $L_{s+1}$ is a Lipschitz constant. Therefore,
\begin{align}
    \nonumber\mathcal{C}_{s+1}=&\arg\min_{\mathcal{C}}\frac{1}{2}||\mathcal{C}-(\mathcal{C}_s-\frac{1}{L_{s+1}}\nabla f(\mathcal{C}_s))||^2_F\\
    &+ \frac{\lambda}{L_{s+1}}||\mathcal{C}||_1,
\end{align}
We can obtain the Lipschitz constant that $L=\sum^n_{b=1}||\Tilde{\mathcal{D}}^{{(b)}^H}\Tilde{\mathcal{D}}^{(b)}||^2_F$, $\Tilde{\mathcal{D}}^{(b)}$ is the discrete fourier transformation (DFT) of the third-dimension slice ${\mathcal{D}}^{(b)}(:,j),b=1,2,...n$, and subscript $H$ implies that it is a conjugate transpose. In the implemented algorithm for the training process of $\mathcal{C}$, we use $\mathbf{Prox}_{\beta / L}$ to solve above equations, which is the proximal operator~\cite{parikh2014proximal}. We therefore obtain the tensor sparse coding coefficients $\mathcal{C}$ through iteratively solving Equ. (12).

For learning the dictionary $\mathcal{D}$ with fixed $\mathcal{C}$, the optimization problem w.r.t each of the $n$ slices of $\mathcal{D}$ becomes
\begin{align}
 &\min_{\mathcal{D}^{(b)}\in\mathbb{R}^{d\times m}, b=1,2...n}  ||{\mathcal{X}}^{(b)}-{\mathcal{D}}^{(b)}\ast{\mathcal{C}}^{(b)}||^2_F
    \\
    &\textrm{s.t.}  ||\Tilde{\mathcal{D}}^{(b)}(:,j)||^2_F\leqslant1, j=1,2,...,m, b=1,2,...,n.
\end{align}
Transform the above equations into the frequency domain,
\begin{equation}
    \min_{\mathcal{D}^{(b)}\in\mathbb{R}^{d\times m}, b=1,2...n}||\Tilde{\mathcal{X}}^{(b)}-\Tilde{\mathcal{D}}^{(b)}\ast\Tilde{\mathcal{C}}^{(b)}||^2_F
\end{equation}
\begin{equation}
    \textrm{s.t.} ||\Tilde{\mathcal{D}}^{(b)}(:,j)||^2_F\leqslant1, j=1,2,...,m, b=1,2,...,n.
\end{equation}
Therefore, with the Langrange dual, we obtain
\begin{align}
   \nonumber\mathcal{L}(\Tilde{\mathcal{D}},\Omega)&=\sum_{b=1}^n  ||\Tilde{\mathcal{X}}^{(b)}-\Tilde{\mathcal{D}}^{(b)}\ast\Tilde{\mathcal{C}}^{(b)}||^2_F+\\
 &\sum_{j=1}^m\omega_j(\sum^n_{b=1}||\Tilde{\mathcal{D}}^{(b)}(:,j)||^2_F-n).
\end{align}
Thus, the optimal formulation of $\widehat{\mathcal{D}}^{(b)}$ satisfies:
\begin{equation}
    \Tilde{\mathcal{D}}^{(b)}=(\Tilde{\mathcal{X}}^{(b)}\Tilde{\mathcal{C}}^{{(b)}^H})(\Tilde{\mathcal{Z}}^{(b)}\Tilde{\mathcal{C}}^{{(b)}^H}+\Omega)^{-1}.
\end{equation}
Therefore,
\begin{equation}
    \mathcal{L}(\Omega)=-\sum^n_{b=1}\textbf{Tr}(\Tilde{\mathcal{C}}^{{(b)}^H}\Tilde{\mathcal{X}}^{(b)}\Tilde{\mathcal{D}}^{{(b)}^H})-n\sum_{j=1}^m\omega_j.
\end{equation}
Equ. (19) can be solved with Newton's method. Substitute the derived $\Omega$ in Equ. (18). Thus, we can derive the dictionary $\mathcal{D}$ through inverse fourier transformation of $\Tilde{\mathcal{D}}^{(b)}$.
\subsection{The Generation Phase}
In the generation phase, we first generate low-resolution images ${T}_G\in \mathbb{R}^{p\times q}$ with the trained DCGAN model directly from latent vectors $\textbf{r}$ in random distribution, and concatenate them to make image tensors $\mathcal{T}_G\in \mathbb{R}^{p\times q\times r}$. Then, we set $\mathcal{T}_l=\mathcal{T}_G$ to derive the tensor sparse coefficients $\mathcal{C}$ with the relationship $\mathcal{T}_l = \mathcal{D}_l * \mathcal{C}$ and trained dictionary $\mathcal{D}_l$ with $\mathcal{C}$ (here the ``trained'' dictionary does not mean the dictionary is derived through a training process like the neural networks, but a specific iteration algorithm for deriving the dictionary, see details in Section 4.3 and 4.4. Finally we use $\mathcal{T}^{\prime}_h = \mathcal{D}_h * \mathcal{C}$ to generate high-resolution output tensor block $\mathcal{T}^{\prime}_h$ with derived dictionary $\mathcal{D}_h$. The output high-resolution 2D images $X^\prime \in \mathbb{R}^{p\times q}$ are obtained through ``unfolding'' the generated high-resolution tensor block $\mathcal{T}^{\prime}_h$.


\begin{algorithm}[htbp]
\caption{Deep Tensor Generative Adversarial Net (TGAN) - Training Phase}
\label{algorithm: TGAN1}
\begin{algorithmic}[1]
\STATE {\bf Input:} original images $X\in \mathbb{R}^{p\times q}$, training iteration $T,S$, sparsity parameter $\lambda$;
\STATE {\bf Initialization:} high-resolution and low-resolution tensor dictionaries $\mathcal{D}_h\in\mathbb{R}^{d_h\times m \times n},\mathcal{D}_l\in\mathbb{R}^{d_l\times m \times n}$, common coefficients $\mathcal{C}_0:=\mathbf{0}\in \mathbb{R}^{m\times N\times n}$ ($N$ is the number of samples used for training the dictionaries), and Lagrange dual variables $\omega \in \mathbb{R}$, $\mathcal{B}_1=\mathcal{C}_0$, $t_1=1$;
\STATE Concatenate $r$ different-direction pixel-shifting images from the same original image $X$ to be high-resolution image tensors $\mathcal{X}_h \in \mathbb{R}^{p\times q\times r}$, and downsample it at the downsampling rate $c$ to be low-resolution image tensors $\mathcal{X}_l\in\mathbb{R}^{\frac{p}{c}\times\frac{q}{c}\times r}$; 
\STATE Sample $\mathcal{X}_h, \mathcal{X}_l$ (both of number $N$) by small tensor cubes with stridesto generate input tensor blocks $\mathcal{T}_h\in \mathbb{R}^{d_h\times N\times n}, \mathcal{T}_l\in \mathbb{R}^{d_l\times N\times n}$ (as is called the sampling and ``folding'' process in Section 4.2);
\STATE Train DCGAN with the ${X}_l$ training set to generate low-resolution images $T_G\in\mathbb{R}^{\frac{p}{c}\times\frac{q}{c}} $  from the latent vector $\textbf{r}\in \mathbb{R}^{u\times 1}$ in random distributions, and use discriminator to distinguish between the generated images and original input ${X}_l$. Update the DCGAN through backpropagation of the mean squared error (MSE) loss. 

\FOR{$k=1$ to $T$}
\STATE  \textit{\# Solve tensor coefficient $\mathcal{C}$}.
\FOR{$s=1$ to $S$}
\STATE{Set $L_s = \eta_s(\sum_{b=1}^n\|\widehat{\mathcal{D}}^{(b)^H}\widehat{\mathcal{D}}^{(b)}\|_F)$;}
     \STATE{Compute $\nabla f(\mathcal{B}_s)$ ;}
     \STATE{Compute $\mathcal{C}_s$ via $\textbf{Prox}_{\beta/L_s}(\mathcal{B}_s-\frac{1}{L_s}\nabla f(\mathcal{B}_s))$;}
     \STATE{$t_{s+1}=\frac{1+\sqrt{1+4t^2_s}}{2}$;}
     \STATE{$\mathcal{B}_{s+1}=\mathcal{C}_s + \frac{t_s-1}{t_{s+1}}(\mathcal{C}_s-\mathcal{C}_{s-1})$;}
 \ENDFOR
\STATE  \textit{\# Solve tensor dictionaries $\mathcal{D}_h, \mathcal{D}_l$}.
\STATE Take Fourier transformation for $\mathcal{T}=[1/\sqrt{N}\mathcal{T}_h, 1/\sqrt{N}\mathcal{T}_l]^T$ to obtain $\Tilde{\mathcal{T}}$ and $\Tilde{\mathcal{C}}$;
\STATE Solve Equ. (19) for $\omega$ via Newton's method;
\STATE Derive $\Tilde{\mathcal{D}}^{(b)}$ from Equ. (18), $l=1,2,...,n$;
\STATE Take inverse Fourier transformation of $\Tilde{\mathcal{D}}$ to derive $\mathcal{D}$. $\mathcal{D}$ includes feature dictionary $\mathcal{D}_l$ and recovery dictionary $\mathcal{D}_h$.
\ENDFOR
\STATE {\bf Output:} feature dictionary $\mathcal{D}_l$ and recovery dictionary $\mathcal{D}_h$.

\end{algorithmic}
\end{algorithm} 

\begin{algorithm}[htbp]
\caption{Deep Tensor Generative Adversarial Net (TGAN) - Generation Phase}
\label{algorithm: TGAN2}
\begin{algorithmic}[1]
\STATE {\bf Input:} $\mathcal{D}_h\in\mathbb{R}^{d_h\times m \times n},\mathcal{D}_l\in\mathbb{R}^{d_l\times m \times n}$;

\STATE Use the trained DCGAN to generate low-resolution images $T_G\in\mathbb{R}^{\frac{p}{c}\times\frac{q}{c}} $  from the latent vector $\textbf{r}\in \mathbb{R}^{u\times 1}$ in random distributions, and further concatenate $r$ images $T_G$ to image tensors $\mathcal{T}_G\in \mathbb{R}^{\frac{p}{c}\times\frac{q}{c}\times r}$ ;

\STATE Reshape the low-resolution image tensor $\mathcal{T}_G$ generated with DCGAN to be $\mathcal{T}^\prime_l\in \mathbb{R}^{d_l\times N^\prime \times n}$ through sampling and ``folding'', and use $\mathcal{T}^\prime_l=\mathcal{D}_l\ast\mathcal{C}$ to obtain tensor sparse coding coefficients $\mathcal{C}$ with feature dictionary $\mathcal{D}_l$ derived above;
\STATE Use $\mathcal{T}^\prime_h=\mathcal{D}_h\ast\mathcal{C}$ to generate high-resolution tensor images $\mathcal{X}_h$ with tensor sparse coding coefficients $\mathcal{C}$ and recovery dictionary $\mathcal{D}_h$;
\STATE Transform high-resolution tensor images $\mathcal{T}^\prime_h$ into 2D images ${X'}\in \mathbb{R}^{p\times q}$ (through the ``unfolding'' process);
\STATE {\bf Output:} Generated high-resolution 2D images ${X'}$.
\end{algorithmic}
\end{algorithm}

\section{Performance Evaluation}
\label{sect: Performance Evaluation}
In this section, we present the results of proposed TGAN scheme on three datasets: MNIST~\cite{lecun1998gradient}, CIFAR10~\cite{krizhevsky2009learning}, PASCAL2 VOC~\cite{Everingham:2010:PVO:1747084.1747104}. The image size of these three datasets applied in our model is $28\times 28, 32\times 32, 374\times 374$ (downscaled from original $375\times 500$ pixels), repectively.

\subsection{Experiments Setting}
DCGAN neural network parameters: the generator network has one fully connected layer and three transposed convolutional layers, with  a  decreasing  number  of $5\times 5$ filter kernels, decreasing by a factor of 2 from $4\times 64$ to 64 kernels and finally one channel of output images. The discriminator has three convolutional layers, with  an  increasing  number  of $5\times 5$ filter kernels consistent with the generator.
We use LeakyReLu~\cite{xu2015empirical} with parameter $\alpha=0.2$ to avoid max-pooling.
Strided convolutions of size $[1,2,2,1]$ are used in each convolutional layer and tranposed convolutional layer. The learning rate is set to $1\times 10^{-4}$ and stochastic gradient descent is applied with
a mini-batch size of 32. 

By default, $u=128, d_h=16, d_l=96, m=128, n=4, N =10000, N^\prime =2500$. The number of directions for pixel-shifting is $r=7$. The number of iterations $T=10, S=50$. The sparsity parameter $\lambda=0.05$. $\beta$ in $\textbf{Prox}$ method is 0.05. For MNIST data, original images of size $p\times q, p=28, q=28$ (size values are set accordingly for other two datasets), downsampling rate of low-resolution images compared with high-resolution images is $c=2$.


\begin{figure}[t]
    \centering
    \includegraphics[height=2.2cm]{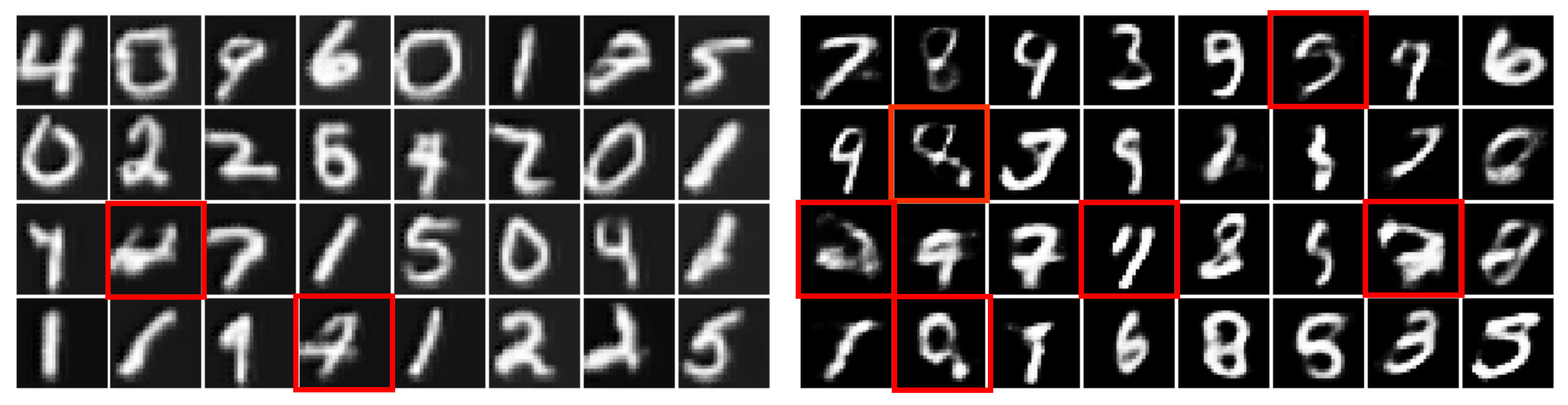}
     \caption{MNIST samples of $28 \times 28$ pixels: for TGAN and AAE model, we pick the generated digital number images which are hard to recognize (in red borders). The number of the obscure images of TGAN (left) and AAE (right) is 2 and 6, respectively. }
  \label{figure: generated samples}
\end{figure}

\begin{figure}[htbp]
    \centering
    \includegraphics[height=4cm]{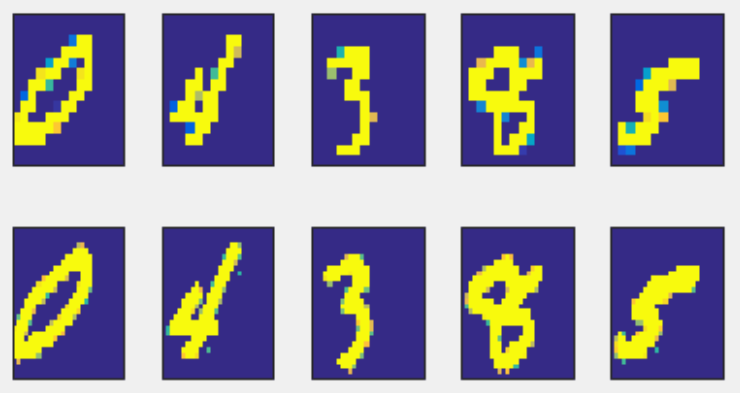}
     \caption{Ablation studies: MNIST samples using TGAN with (below) or without (above) tensor super-resolution. This testifies the significant effects of tensor-based super-resolution process.}
  \label{figure: CIFAR10 samples}
\end{figure}

\subsection{Inception Score of Generation Results}
We adopt the inception score (IS) metric \cite{salimans2016improved}\cite{szegedy2016rethinking} to compare performance of different schemes. The metric compares three kinds of samples, including our generated images, other generated images from similar generative methods and the real images. The inception score metric focus on comparing the qualities and diversities of their generated images. 
We input every generated image in Google Inception Net and obtained the conditional label distribution $p(y|\textbf{x})$, where $\textbf{x}$ is one generated image and y denotes the predicted label. Images that contain meaningful objects should have a conditional label distribution with low entropy. 
The inception score metric is $\text{exp} \left[\mathbb{E}_{\textbf{x} \sim X^\prime}\mathbb{KL}(p(y|\textbf{x})||p(y))\right]$. The comparison results of the AAE and our TGAN model are shown in Table \ref{table: LLH}. The proposed TGAN achieves better results in all three datasets, especially for larges-sized PASCAL2 images (e.g. $374\times 374$). Its inception score of 4.02 for PASCAL2 images significantly outperforms AAE of 3.81.

\begin{table}[h]
  \centering
  \begin{tabular}{c|c|c}
    Dataset     &CIFAR 10 & Pascal2 VOC \\
    \midrule
   \centering  AAE~\cite{makhzani2015adversarial}     & 3.98 & 3.81 \\
    \centering TGAN     & $\mathbf{4.05}$ & $\mathbf{4.02}$  \\
    \bottomrule
  \end{tabular}
  \caption{The inception score estimates metric are measured for AAE and our proposed TGAN model on CIFAR10 and Pascal2 VOC datasets.}
  \label{table: LLH}
\end{table}

\subsection{Generated Images of TGAN}
Some of the testing results on benchmark datasets are shown in the end of the paper. Fig. 3 shows the comparison of MNIST images generation with TGAN and AAE. In the random selected 16 images, only 2 of TGAN generated images is kind of obscure to recognize, compared with at least 6 in AAE generated ones. The TGAN model provides images with more precise features of digital numbers, which benefits from its concise and efficient representation in tensor space.   The effects of tensor super-resolution are shown in Fig. 4 for MINIST images with ablation studies. The images generated with general DCGAN have much coarser features without the tensor-based super-resolution process, which testifies that tensor super-resolution can significantly increases the image quality with more convincing details. Fig. 5 and Fig. 6 shows the generation results on PASCAL2 and CIFAR10 datasets, both testify the capability of TGAN in generating images with better quality, especially for large images (e.g. $374 \times 374$) in PASCAL2. Images generated with TGAN have more precise features and convincing details than images generated by AAE. This testifies that TGAN preserves spatial structure and local proximal information in a better way than traditional methods.  Generally, the DCGAN generates basic shapes, structures, and colors of images, while the cascading tensor super-resolution process improves the images with more details.

\begin{figure*}[t]
    \centering
    \includegraphics[height=6cm]{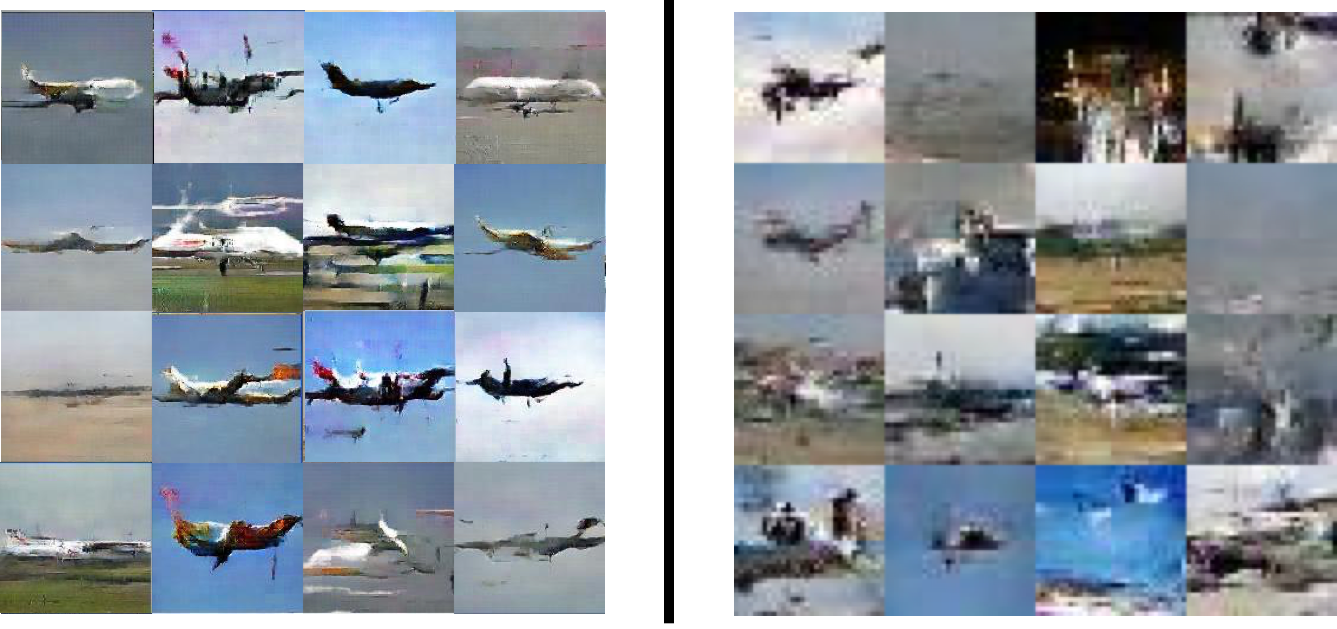}
     \caption{PASCAL2 samples of $374 \times 374$ pixels: we show the large size airplane samples generated by TGAN, compared with the same-sized samples generated by AAE for airplane images in PASCAL2.}
  \label{figure: CIFAR10 samples}
\end{figure*}

\begin{figure}[htbp]
    \centering
    \includegraphics[height=8cm]{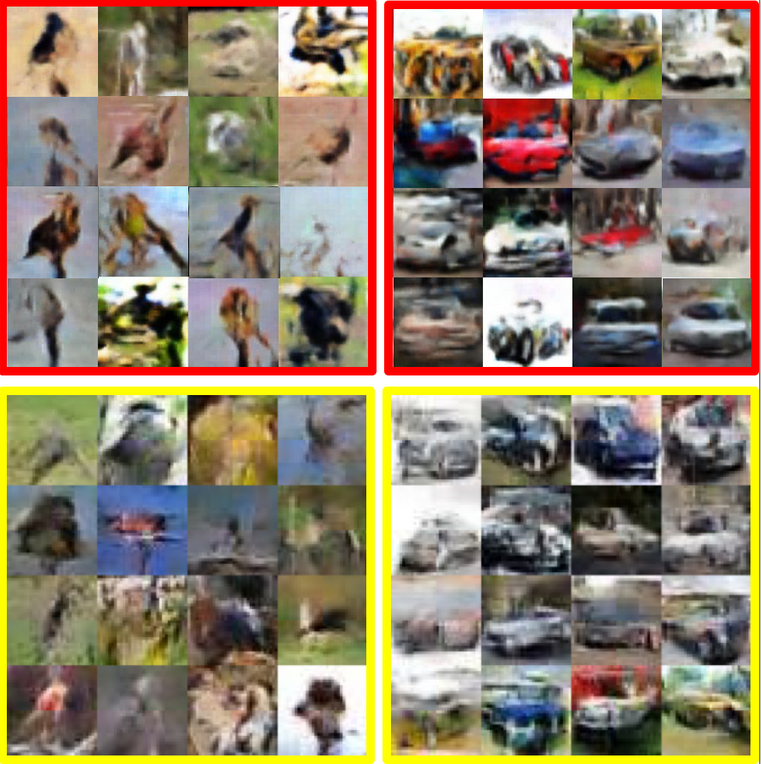}
     \caption{CIFAR10 samples of $128 \times 128$ pixels ($4\times4$ image matrix of $32\times 32$ pixels images ): TGAN and AAE model. We show three kinds of samples: airplane, bird, and car. The pictures with red borders are generated by TGAN, while pictures with yellow borders are generated by the AAE model.}
  \label{figure: generated samples}
\end{figure}

\section{Conclusion}
\label{sect: Conclusion}

In this paper, we proposed a TGAN scheme that integrates DCGAN model and tensor super-resolution, which is able to generate large-sized high-quality images. The proposed scheme applies tensor representation space as main operation space for image generation, which shows better results than traditional generative models working in image pixel space. Essentially, the adversarial process of TGAN takes place in a tensor space. Note that in the tensor super-resolution process, tensor sparse coding brings several advantages: (i) the size of dictionary, which accelerates the training process for deriving the representation dictionary; (ii) more concise and efficient representation for images, which is verified in the generated images in our experiments. TGAN is superior in preserving spatial structures and local proximity information in images. Accordingly, the tensor super-resolution benefits from tensor representation to generate higher-quality images, especially for large images. Our proposed cascading TGAN scheme surpasses the state-of-the-art generative model AAE on three datasets (MNIST, CIFAR10, and PASCAL2). 




\newpage

\small

\bibliographystyle{IEEEbib}
\bibliography{ref}

\end{document}